\pdfoutput=1

\documentclass[11pt]{article}

\usepackage[]{EMNLP2023}

\usepackage{amsmath,amsfonts,bm}

\def\eqref#1{equation~\ref{#1}}

\def\1{\bm{1}}

\DeclareMathAlphabet{\mathsfit}{\encodingdefault}{\sfdefault}{m}{sl}
\SetMathAlphabet{\mathsfit}{bold}{\encodingdefault}{\sfdefault}{bx}{n}

\usepackage{wrapfig,lipsum,booktabs}
\usepackage{hyperref}
\usepackage{url}
\usepackage{graphicx}
\usepackage{amsmath,amssymb,amsfonts}
\usepackage{pifont}
\usepackage{array}
\usepackage{multirow}
\usepackage[export]{adjustbox}
\usepackage{caption}
\usepackage{subcaption}
\usepackage{float}
\usepackage{newfloat}
\usepackage{makecell}
\usepackage{bbm}
\usepackage{bm}	
\usepackage{subcaption}
\usepackage{placeins}
\usepackage{listings}
\usepackage{bibentry}
\usepackage{arydshln}
\usepackage[noend]{algpseudocode}

\usepackage{times}
\usepackage{latexsym}

\usepackage[T1]{fontenc}

\usepackage[utf8]{inputenc}

\usepackage{microtype}

\usepackage{inconsolata}

\title{Merging Experts into One: \\ Improving Computational Efficiency of Mixture of Experts}

\newcommand*\samethanks[1][\value{footnote}]{\footnotemark[#1]}

\author{Shwai He\textsuperscript{\rm 1}\space\space
Run-Ze Fan\textsuperscript{\rm 3}\space\space
Liang Ding\textsuperscript{\rm 2}\thanks{~~Corresponding author}\space\space
Li Shen\textsuperscript{\rm 4}\space\space 
Tianyi Zhou\textsuperscript{\rm 1}\samethanks\space\space 
Dacheng Tao\textsuperscript{\rm 2}\\
    \textsuperscript{\rm 1}University of Maryland, College Park \space\space
    \textsuperscript{\rm 2}The University of Sydney\\
    \textsuperscript{\rm 3}University of Chinese Academy of Sciences \space\space
    \textsuperscript{\rm 4}JD Explore Academy\\
    {\tt\small shwaihe@umd.edu},\space\space
    {\tt\small liangding.liam@gmail.com},\space\space
    {\tt\small tianyi@umd.edu}\space\space
}

\makeatletter
\newcommand{\biggg}{\bBigg@{1.2}}

\makeatother

\newcommand{\ignore}[1]{{}}

\begin{document}

\maketitle

\begin{abstract}
Scaling the size of language models usually leads to remarkable advancements in NLP tasks. But it often comes with a price of growing computational cost. Although a sparse Mixture of Experts (MoE) can reduce the cost by activating a small subset of parameters (e.g., one expert) for each input, its computation escalates significantly if increasing the number of activated experts, limiting its practical utility. Can we retain the advantages of adding more experts without substantially increasing the computational costs? In this paper, we first demonstrate the superiority of selecting multiple experts and then propose a computation-efficient approach called \textbf{\texttt{Merging Experts into One}} (MEO), which reduces the computation cost to that of a single expert. Extensive experiments show that MEO significantly improves computational efficiency, e.g., FLOPS drops from 72.0G of vanilla MoE to 28.9G (MEO). Moreover, we propose a token-level attention block that further enhances the efficiency and performance of token-level MEO, e.g., 83.3\% (MEO) vs. 82.6\% (vanilla MoE) average score on the GLUE benchmark. Our code will be released upon acceptance. Code will be released at: \url{https://github.com/Shwai-He/MEO}.

\end{abstract}

\section{Introduction}
\label{sec:introduction}

Scaling language models has achieved promising progress in the field of NLP~\cite{DBLP:conf/nips/BrownMRSKDNSSAA20, DBLP:journals/corr/abs-2303-08774}. To further increase the model size under a computational budget, sparsely activated networks~\cite{DBLP:conf/icml/DuHDTLXKZYFZFBZ22, DBLP:conf/emnlp/ArtetxeBGMOSLDI22} only employ a few parameters for each input. 
A widely studied approach is the Mixture-of-Experts~\citealp[(MoE,][]{DBLP:conf/iclr/ShazeerMMDLHD17}), which trains multiple expert networks but only selects a subset of them for a specific input~\cite{DBLP:journals/neco/JacobsJNH91, DBLP:journals/neco/JordanJ94}. Compared to dense networks of the same model size, MoE effectively reduces computational costs. \looseness-1

\begin{figure}[t]
\centering
\makeatother\def\@captype{figure}\makeatother
\centering
    \hspace{-8pt}    
\includegraphics[width=0.48\textwidth]{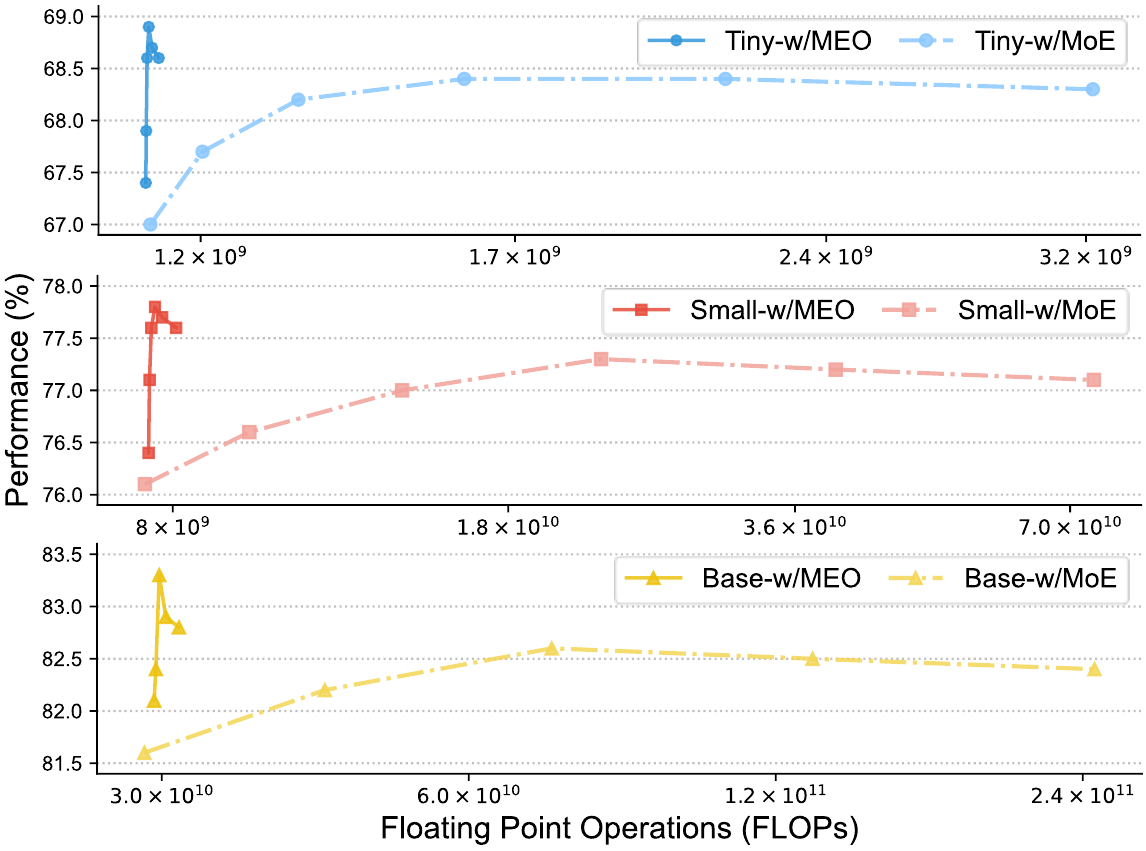}
 \caption{\textbf{Performance vs. FLOPs of MoE and MEO} at the token level when different numbers of experts (i.e., 1, 2, 4, 8, 16, 32) are selected. We take three different sizes of BERT as the expert model. \looseness-1}
 \vspace{-10pt}
\label{fig:Intro}
\end{figure}

Although increasing the experts selected for each input can improve the representation diversity~\cite{DBLP:conf/nips/YangBLN19} and downstream task performance~\cite{DBLP:conf/iclr/ShazeerMMDLHD17, DBLP:conf/nips/YangBLN19}, it usually comes with a price of significantly growing computational cost. 
Our empirical study (Figure~\ref{fig:Intro} and Table~\ref{tab:pre_exp}) verifies the Pros and Cons (\textit{superior performance} vs. \textit{high computational cost}) of selecting multiple experts at MoE inference.
Hence, to retain the advantage of MoE on computational efficiency, existing work mainly selects only one expert per input in applications or experiments~\cite{fedus2021switch}, which inevitably compromises the performance.

\begin{figure*}[htbp]
\centering
\makeatother\def\@captype{figure}\makeatother
	\centering
\begin{subfigure}[h]{0.48\linewidth}
    \centering
\includegraphics[width=\linewidth]{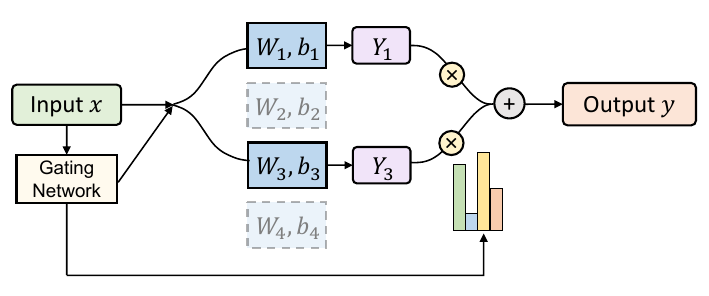}
    \subcaption{MoE}
\end{subfigure}\hspace{10pt}\begin{subfigure}[h]{0.48\linewidth}
    \centering
    \includegraphics[width=\linewidth]{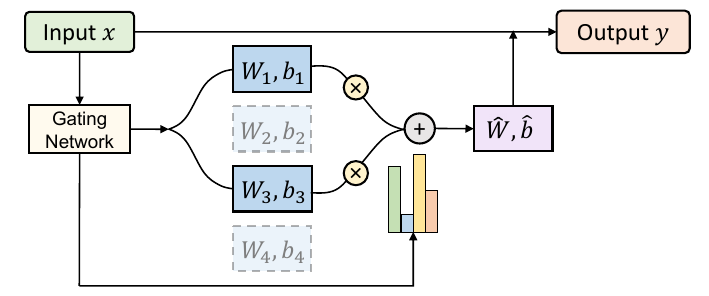}
    \subcaption{MEO}
\end{subfigure}
\caption{\textbf{The diagrams of (a) MoE and (b) our proposed MEO}, with a case of $m=2$ experts are selected. MoE linearly combines the outputs from experts, while MEO first merges experts into one and then computes the input. } 
\label{fig:overall_method}
\vspace{-10pt}
\end{figure*}

Our work aims to improve the computational efficiency of MoE inference with multiple experts selected, for greatly rejuvenating the compromised performance.
The computation involved in MoE primarily consists of the inference on each selected expert and the summation of their outputs, with the former dominating the cost. Hence, the cost linearly grows with the number of selected experts. 
To overcome the computational bottleneck, we instead propose \textbf{Merging Experts into One} (MEO), which alters the calculation order of the two operations, i.e., first merging the parameters of the selected experts into one expert followed by inference on the merged expert. Since the parameter merging only requires summation, an MEO layer (approximately) only consumes the computation of single-expert inference, no matter how many experts are selected. 
This leads to a nearly constant inference cost when scaling up the model capacity (i.e., the number of selected experts) to improve the performance. 
 
MEO can be applied as a drop-in replacement for MoE, which has been deployed at various levels, e.g., selecting experts for each token~\cite{DBLP:conf/iclr/ShazeerMMDLHD17}, each sequence~\cite{ye-2022-eliciting}, each task~\cite{DBLP:conf/emnlp/KuduguntaHBKLLF21}, etc. On the sequence/task level, our empirical studies demonstrate that replacing MoE with MEO significantly improves computational efficiency, e.g., reducing FLOPs from 72.0G to 28.9G, without hurting the performance. In addition, we propose a token-level attention mechanism that further enhances the efficiency and performance, e.g., from 82.6\% (MoE) to 83.3\% (MEO) on BERT-Base (Figure~\ref{fig:Intro}).

\section{Methodology}
\label{sec:method}

\paragraph{Review of Mixture of Experts.}
Given a token $x_i$ in the input sequence $x \in \mathbb{R}^{s \times d}$, MoE selects $m$ experts from $n$ $(m \leq n)$ experts $(E_1, \dots, E_n)$ based on a gating network. We denote $\mathcal{G}$ as the gating scores and $\mathcal{T}$ as the indices of selected experts. MoE linearly combines the outputs of selected experts: 
\begin{equation}
y_i = \sum\nolimits_{k \in \mathcal T} \mathcal{G}_k(x_i) \cdot E_k(x_i).
\label{eq:moe}
\end{equation}
MoE performs at various levels, e.g., token, sequence, and task,  where MoE selects experts based on a single token, input sequence, or task embedding (or task ids):
\begin{equation}
\small
 \mathcal{G}(x_i) = 
 \begin{cases}
 \text{GATE}(x_i), & \text {Token-level} \\
 \text{GATE}(\frac{1}{s}\sum_{i=1}^s x_i), & \text {Sequence-level} \\
 \text{GATE}(task\_ids), & \text {Task-level}
 \end{cases}, 
\end{equation}
where ``GATE'' denotes the gating function. 

\begin{table}[ht]
    \centering
    \caption{\textbf{Effects of the number of selected experts} on performance. The best results are \textbf{bold}. }
    \resizebox{\columnwidth}{!}{
    \setlength{\tabcolsep}{2pt}
    \begin{tabular}{ccccccc}
    \toprule
     \bf ~$m$~ & ~\#FLOPs.~ & ~SST-2~ & ~STSB~ & ~MNLI~ & ~QNLI~ & ~~\underline{Avg.}~ \\
     \midrule
        1~ 
    & 7.5G & 87.1 & 86.1 & 77.8 & 85.8 & \underline{84.2} \\
        2~ 
    & 9.6G & 87.9 & 86.8 & 78.2 & 86.2 & \underline{84.8} \\
        4~ 
    & 13.9G & 88.2 & 87.1 & 78.3 & 86.4 & \underline{85.0} \\
        8~ 
    & 22.5G & 88.3 & \bf 87.7 & \bf 79.1 & \bf 86.8 & \bf \underline{85.5} \\
        16~ 
    & 39.7G & \bf 88.4 & 87.5 & 78.8 & 86.6 & \underline{85.3} \\
        32~ 
    & 74.1G & 88.2 & 87.6 & 78.6 & 86.3 & \underline{85.2} \\
    \bottomrule
    \end{tabular}}
\vspace{-10pt}
\label{tab:pre_exp}
\end{table}

\paragraph{Motivation.}
While many predominant MoE models tend to select the top-$1$ expert \cite{fedus2021switch}, 
selecting multiple experts has the potential of boosting the representation power \cite{DBLP:conf/cvpr/ChenDLCYL20, DBLP:conf/nips/YangBLN19}. Empirically, we conduct preliminary experiments on the BERT-Small \cite{bhargava-2021-generalization} to verify it.

In Table \ref{tab:pre_exp}, it is evident that selecting multiple experts contributes to better performance. Even though selecting excessive experts is suboptimal as it introduces the interference between experts that hinders the performance \cite{DBLP:conf/nips/MustafaRPJH22, DBLP:conf/nips/ZhuZWWLWD22}, our preliminary experiments necessitates the selection of multiple experts. 

However, selecting more experts leads to a substantial increase in FLOPs (e.g., 74.1G v.s. 7.5G when increasing $m$ from 1 to 32). This phenomenon urges us to reflect \textbf{\textit{whether there exists an efficient approach to achieve both high performance and computational efficiency}}. Our goal is to {ensure consistent computational cost, regardless of the number of selected experts.}

\paragraph{Merging Experts into One.}
The computation cost of MoE primarily involves the computation of individual experts (i.e., $\sum_{k \in \mathcal{T}} O(E_k)$) and the mixture of outputs from experts (i.e., $O(\mathcal{G})$ and $O(\sum_{k \in \mathcal{T}}\mathcal{G}_k \cdot E_k)$). Notably, the computation of individual experts plays a dominant role, with even the cost of a single expert being significantly outweighing that of the mixture: 
\begin{equation}
O(E_k) \gg O(\mathcal{G}) + O(\sum_{k \in \mathcal{T}}\mathcal{G}_k \cdot E_k), 
\end{equation}
where $O(\cdot)$ measures the computational cost. 

On the other hand, as the number of selected experts $m$ increases, the term $\sum_{k \in \mathcal{T}} O(E_k)$ experiences a substantial increase, whereas the increase in $O(\sum_{k \in \mathcal{T}}\mathcal{G}_k \cdot {E}_k)$ is marginal. 
Therefore, it is essential to address the growing trend of $\sum_{k \in \mathcal{T}} O(E_k)$ to enhance computational efficiency. 

As illustrated in Figure \ref{fig:overall_method}, we propose the method called Merging Experts into One (MEO), where the key idea is to leverage the gating scores to aggregate the parameters of the selected experts (which is akin to the simple weighted model fusion mechanism~\cite{li2023deep}):
\begin{equation}
\hat W_i = \sum_{k \in \mathcal{T}}  \mathcal{G}_k(x_i) \cdot W_k, \hat b_i = \sum_{k \in \mathcal{T}}  \mathcal{G}_k(x_i) \cdot b_k, 
\end{equation}
where $W_k, b_k$ represent the weight and bias of the $k$-th expert, while $\hat W_i, \hat b_i$ are the aggregated weight and bias for $x_i$. The output of MEO is given by: 
\begin{equation}
y_i = \sigma(\hat W_i x_i + \hat b_i), 
\end{equation}
where $\sigma$ represents the activation function. 

The computation cost of MEO primarily consists of $O(\sigma(\hat W_i x_i + \hat b_i))$, $O(\sum_{k \in \mathcal{T}}\mathcal{G}_k \cdot W_k)$, $O(\sum_{k \in \mathcal{T}}\mathcal{G}_k \cdot b_k)$, and $O(\mathcal{G})$. Among them, $O(\sigma(\hat W_i x_i + \hat b_i))$ is the dominant factor. It is worth noting that $O(\sigma(\hat W_i x_i + \hat b_i))$ is equivalent to the computation cost of a fully connected network and independent of the number of selected experts. Therefore, MEO compresses computation costs significantly.

\paragraph{MEO at Different Levels.}
In the case of sequence and task level MEO, all tokens within a sequence share the same gating scores, as well as the aggregated parameters $\hat{W}$ and $\hat b$ \footnote{we omit subscripts of $\hat{W}$ and $\hat b$ at the sequence and task level given each token shares the same aggregated parameters.}. This property allows for easy adoption of MEO at these levels.

However, when directly applying MEO at the token level,  the situation is different. Since the gating scores of each token within a sequence are unique, the straightforward usage of MEO would require the aggregation of multiple sets of weights and biases, resulting in increased deployment cost. Therefore, we refine and enhance the framework of token-level MEO specifically.

\paragraph{Token-Level MEO.}
Our proposed token-level MEO aims to incorporate token-level information with minimal extra computational cost. Specifically, the expert selection is performed at the sequence level, thereby preserving context information and eliminating the necessity of aggregating multiple weights and biases for individual tokens. To capture the identification of each token, we leverage the token attention mechanism inspired by \citet{houlsby2019parameter, li2021omni}. 

Specifically, given the input sequence $x \in \mathbb{R}^{s \times d}$, we employ a specialized bottleneck block, inspired by adapter-like structures \cite{houlsby2019parameter, pfeiffer2020adapterfusion}. The bottleneck layer incorporates down-projection weights $\boldsymbol{W}_{down}  \in \mathbb{R}^{d \times \frac{d}{r}}$, an activation function $f$ and up-projection weights $\boldsymbol{W}_{up} \in \mathbb{R}^{\frac{d}{r} \times d}$, with reduce factor $r=64$ that ensures low extra computational cost. By operating on each token individually, the bottleneck applies token-level attention to the input sequence $x$: 
\begin{equation}
     x \leftarrow x + f(x \boldsymbol{W}_{down})\boldsymbol{W}_{up}. 
\end{equation}
With the inclusion of token identification in the updated input, MEO performs aggregation of $\hat W$ and $\hat b$ through sequence-level expert selection. Subsequently, these aggregated $\hat W$ and $\hat b$ are used to compute the output in conjunction with the input. 

\begin{table*}
\vspace{-6pt}
\centering
\caption{\textbf{Empirical results for MEO and MoE in task-level ($task$) and sequence-level ($seq$)}. We also report the performance of vanilla feedforward layers (``Vanilla'') as a reference. The shown results are the averaged score for 5 runs. The best results are \textbf{bold}. \ding{86} indicates the method with the fewest the fewer FLOPs (``Vanilla'' is not included). }
\vspace{-5pt}
\resizebox{\linewidth}{!}{ 
\begin{tabular}{lrccccccccccc}
\toprule
\bf Method~~ & ~~~~\#FLOPs. & ~~CoLA~~ & ~~SST-2~~ & ~~MRPC~~ & ~~STS-B~~ & ~~QQP~~  & ~~MNLI~~  & ~~QNLI~~ & ~~RTE~~  & ~~\underline{Avg}~~  \\
\midrule
Vanilla & 28.5G & 54.6 & 91.1 & 84.6 & 85.8 & 90.2 & 80.6 & 90.4 & 66.4 & \underline{80.5} \\
\hdashline
$\mathrm{MoE}_{task}$ & 72.0G & 58.5 & 91.3 & 85.8 & 89.2 & 90.5 & 82.7 & 90.5 & 69.3 & \underline{82.2} \\
$\mathrm{MEO}_{task}$ & \ding{86}28.9G & 59.1 & 91.2 & 85.5 & 89.3 & 90.4 & 83.0 & 90.9 & 68.9 & \underline{82.3} \\
$\mathrm{MoE}_{seq}$ & 72.0G & 59.8 & 91.5 & \bf 86.5 & \bf 89.5 & 90.6 & 83.4 & 90.7 & \bf 70.4 & \underline{82.8} \\
$\mathrm{MEO}_{seq}$ & \ding{86}28.9G & \bf 60.1 & \bf 91.9 & 86.3 & 89.4 & \bf 90.7 & \bf 83.7 & \bf 91.2 & 70.3 & \bf \underline{83.0} \\
\bottomrule
\end{tabular}
}
\label{tab:main_results}
\end{table*}

\section{Empirical Evaluation}
\label{sec:experiments}

\paragraph{Experimental Setup.}
Experiments were conducted on Four widely-used benchmarks, spanning understanding and generation tasks: (1) GLUE~\cite{wang2018glue}, containing understanding tasks like natural language inference, sentiment analysis, and sentence similarity evaluation; (2) XSum~\cite{narayanetal-2018-dont}, a summarization dataset where the models are required to generate a short summary for a given article; (3) WikiText-2 \cite{merity2016pointer}, a collection of over 100 million tokens extracted from the set of verified Good and Featured articles on Wikipedia where the models are utilized to generate the next tokens; (4) SQuAD v1.1~\cite{DBLP:conf/emnlp/RajpurkarZLL16}, a pair-wise dataset for questions and Wikipedia paragraphs where models select the answer span to the question from the paragraph. 

We follow~\citet{zhong2022e2s2,zhong2022improving,he-etal-2023-pad} to conduct experiments on the widely-used GLUE benchmark, containing understanding tasks like natural language inference, sentiment analysis, sentence similarity evaluation, etc. 
We use Adam~\cite{kingma2014adam} as the optimizer with $\beta_1$, $\beta_2$ = 0.9, 0.98. For regularization, we set the weight decay as 0.1 and grid-search the learning rate from \{1e-5, 5e-5, 1e-4, 5e-4\}, where we warm up the learning rate in the first 10\% steps (of the total training steps). For different data scales, we grid-search the training epoch and batch size from \{5, 10, 15, 20\}, and \{8, 16, 32, 64\}, respectively. The maximum length is 128 for GLUE, 1024 for WikiText, and 384 for SQuAD. For XSum, we set the max length of source articles to be 512 and the max length of the target summary to be 128. We follow previous works~\cite{phang2018sentence, lee2019mixout, dodge2020fine, DBLP:conf/emnlp/wangcort22, he2023mera} to fine-tune the pretrained language models, e.g. BERT~\cite{devlin2018bert}, on the downstream training set and report results using the last checkpoint. 

\paragraph{Main Results.}
Following \citet{DBLP:conf/iclr/ShazeerMMDLHD17, gao-2022-parameter}, we conduct experiments on BERT-Base \cite{devlin2018bert} and replace feed-forward layers (``Vanilla'') with MoE or MEO, with the setting $m=4$ and $n=16$. 
In Table \ref{tab:main_results}, we carefully compare our proposed MEO with MoE at task and sequence levels, in terms of computational efficiency and performance. Compared to MoE, MEO significantly reduces the computation cost while achieving comparable performance. Specifically, compared to vanilla feed-forward layers, the Floating Point Operations (FLOPs) of MEO only increase marginally (i.e., about 1\%), while MoE multiplies the FLOPs about 2.53 times.  

\paragraph{Analysis of Reduced Computation. }
\label{sec:reduced_cost}
Compared to a fully connected layer, MEO only introduces computation in gating network $O(\mathcal{G}(x))$ and merging experts (i.e., $O(\sum_{k \in \mathcal{T}} \mathcal{G}_k \cdot W_k)$ and $O(\sum_{k \in \mathcal{T}} \mathcal{G}_k \cdot b_k)$). The additional computation is minimal compared to that of individual experts. 

\begin{figure}[t]
\centering
\makeatother\def\@captype{figure}\makeatother
	\centering
        \hspace{-8pt}	\includegraphics[width=0.45\textwidth]{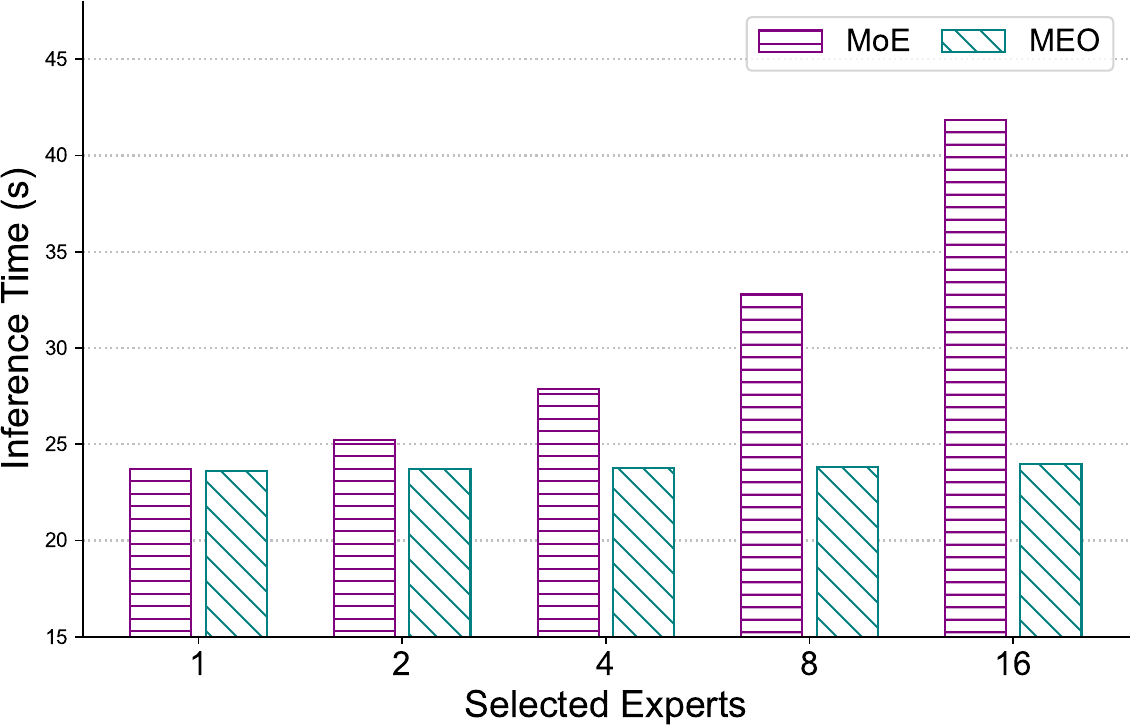}
 \vspace{-5pt}
 \caption{\textbf{Comparison of inference time between MoE and MEO} under a series of different numbers of selected experts (i.e., 1, 2 , 4, 8, 16). }
\label{fig:computation}
\vspace{-8pt}
\end{figure}

In practice, we use eight NVIDIA V100 Tensor Core GPUs to measure the inference time of MEO and MoE on BERT-Base when selecting different numbers of experts (i.e., $n = 1, 2, 4, 8, 16$). Inference time is calculated by the total running time on the MNLI validation dataset with batch size 16. According to Figure \ref{fig:computation}, as the number of selected experts increases, the inference time of MEO is relatively consistent, while MoE exhibits a significantly increased inference time. This highlights the advantage of MEO in computational efficiency, which becomes even more pronounced as the number of selected experts grows.

\begin{table}[ht]
    \centering
    \caption{\textbf{Comparison between MEO and MoE with different activation function usage} (i.e., activation function within (${in}$) and outside (${out}$) experts). }
    \resizebox{\columnwidth}{!}{
    \setlength{\tabcolsep}{2pt}
    \begin{tabular}{lcccccc}
    \toprule
     \bf Method~ & ~FLOPs~ & ~SST-2~ & ~QQP~ & ~MNLI~ & ~QNLI~ & ~~\underline{Avg.}~ \\
     \midrule
        Vanilla~ 
    & 7.5G & 86.9 & 89.1 & 77.2 & 85.2 & \underline{84.6} \\
            $\text{MoE}_{in}$~ 
    & 22.6G & 87.9 & 89.4 & 77.8 & 85.7 & \underline{85.2} \\
        $\text{MoE}_{out}$~ 
    & 22.5G & 87.6 & 89.2 & 78.0 & 85.6 & \underline{85.1} \\
        MEO~ 
    & \ding{86}7.7G & \bf 88.1 & \bf 89.7 & \bf 78.2 & \bf 86.2 & \bf \underline{85.6} \\
    \bottomrule
    \end{tabular}}
\label{tab:nonlinear}
\end{table}

\paragraph{Analysis of Activation Function.}
In many cases where an expert $E_i$ represents a linear layer without a nonlinear activation function, the output of MEO ($y_i=\sigma(\hat W_i x_i + \hat b_i)$) is equivalent to that of MoE ($y_i = \sigma(\sum_{k \in \mathcal{T}} \mathcal{G}_k \cdot (W_k x_i + b_k))$). However, if the expert $E_i$ involves an activation function, the output of MoE is $y_i = \sum_{k \in \mathcal{T}} (\mathcal{G}_k \cdot \sigma(W_k x_i + b_k))$, which leads to differences in outputs and potentially in performance. As depicted in Figure \ref{tab:nonlinear}, we compare MEO with MoE with different usage of activation, where we consider two scenarios: activation function within or outside experts. The results demonstrate that the performance gap between the two scenarios is minimal and indicates the effectiveness of MEO in handling expert networks that incorporate activation functions.  
\begin{figure}[t]
\centering
\makeatother\def\@captype{figure}\makeatother
\vspace{8pt}
	\centering        \hspace{-8pt}
\includegraphics[width=0.45\textwidth]{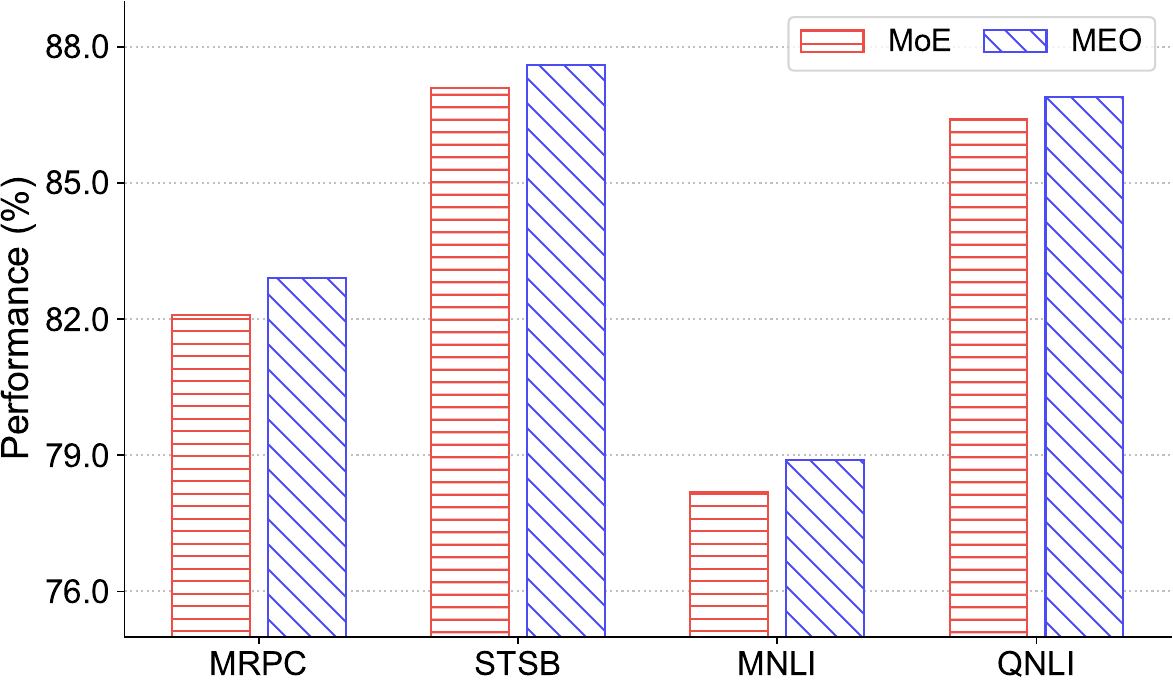}
 \caption{\textbf{Performance comparison between token level MoE and MEO}, where we take BERT-Small as the backbone with the setting $m=8$ and $n=32$. }
\label{fig:token_comparison}
\vspace{-13pt}
\end{figure}

\begin{table}[ht]
    \centering
    \vspace{-5pt}
    \caption{\textbf{Performance of Token-Level MEO}, where we take BERT-Large as the backbone with the setting $m=2$ and $n=8$. }
    \vspace{-5pt}
    \resizebox{\columnwidth}{!}{
    \setlength{\tabcolsep}{2pt}
    \begin{tabular}{lcccccc}
    \toprule
     \bf Method~ & ~FLOPs~ & ~SST-2~ & ~MRPC~ & ~STSB~ & ~QNLI~ & ~~\underline{Avg.}~ \\
     \midrule
        Vanilla~ 
    & 87.2G & 93.2 & 86.8 & 89.1 & 91.8 & \underline{90.2} \\
        MoE~ 
    & 139.0G & 93.7 & 87.2 & 89.7 & 92.2 & \underline{90.7} \\
        MEO~ 
    & \ding{86}91.2G & \bf 94.1 & \bf 87.5 & \bf 89.8 & \bf 92.4 & \bf \underline{91.0} \\
    \bottomrule
    \end{tabular}}
\vspace{-8pt}
\label{tab:large}
\end{table}

\paragraph{Effectiveness of Token-Level MEO.}
For MEO at the token-level MEO, we have incorporated token-level attention blocks. To assess the deployment cost of newly added blocks, we first calculate the extra parameters and FLOPs, with BERT-Small as the backbone. The extra cost of added blocks is minimal (i.e., 0.6M parameters and ~0.15 GFLOPs). Furthermore, in Figure \ref{fig:token_comparison}, we present a performance comparison between token level MEO and MoE in four natural language understanding tasks, where MEO outperforms MoE consistently across these tasks, e.g., 78.9\% v.s. 78.1\% on MNLI. For the average score on the GLUE benchmark, MEO boosts the performance significantly, i.e. 83.3\% v.s. 82.6\% on BERT-Base and 77.8\% v.s. 77.3\% on BERT-Small. 

We also implement the token-level MEO on BERT-Large, utilizing $8$ experts and selecting $2$ experts, resulting in a model with about 1.75 billion parameters. As demonstrated in Table \ref{tab:large}, MEO consistently enhances performance across various tasks, e.g., 0.4\% improvement in SST-2 when compared to MoE. Notably, the additional computational cost is minimal, with only a 4.0 GFLOPs increase over the Vanilla model. Therefore, token-level MEO proves to be an efficient and effective alternative to token-level MoE.

\paragraph{Transfer to different architectures and tasks.}
Utilizing MEO in BERT architectures enhances computational efficiency and performance, and we further validate the effectiveness of MEO on a wide range of architectures for different tasks. In Table \ref{tab:diff}, we use BART-Large \cite{lewisetal2020-bart}  for XSum \cite{narayanetal-2018-dont}, GPT-2-Small \cite{radford2019language} for WikiText \cite{merity2016pointer}, and T5-Base \cite{2020t5} for SQuAD \cite{DBLP:conf/emnlp/RajpurkarZLL16}. MEO and MoE are deployed at the token level. Considering the limited computation resource, we set $m=2$ and $n=8$ for BART and GPT-2, while $m=4$ and $n=16$ are set for T5.  

Clearly, MEO outperforms the standard MoE in three tasks, showing its universality in both natural language understanding and generation.

\begin{table}[h]
    \centering
    \caption{\textbf{Effectiveness on different architectures and tasks.}
    XSum, WikiText, and SQuAD are evaluated with ROUGE-2 (R2.), Perplexity (PPL), and Exact Match (EM), respectively. }
    \resizebox{\columnwidth}{!}{
    \setlength{\tabcolsep}{2pt}
    \begin{tabular}{lcccccc}
    \toprule
    \multirow{2}{*}{\bf Method} & \multicolumn{2}{c}{~XSum}& \multicolumn{2}{c}{~WikiText} & \multicolumn{2}{c}{~SQuAD}\\
    \cmidrule(lr){2-3}
    \cmidrule(lr){4-5}
    \cmidrule(lr){6-7}
    & ~FLOPs~ & ~R2.~ & ~FLOPs~ & ~PPL~ & ~FLOPs~ & ~EM~ \\
    \midrule
    Vanilla & ~369.4G & ~21.9 & ~295.4G & ~21.9 & ~90.2G & ~81.6 \\
    \midrule
    MoE & ~576.6G & ~22.2 & ~412.2G & ~21.1 & ~221.3G & ~82.0 \\
    MEO & ~\ding{86}383.6G & \bf ~22.4 & ~\ding{86}303.2G & \bf ~20.9 & ~\ding{86}93.5G & \bf ~82.1 \\
    \bottomrule
    \end{tabular}}
\label{tab:diff}
\vspace{-8pt}
\end{table}
\section{Conclusion}
\label{sec:conclusion}
In this work, we systematically investigate the computational cost of the Mixture of Experts. Based on our findings, we propose a drop-in replacement called Merging Experts into One (MEO) to enhance computational efficiency.
Additionally, we propose a Token-Level attention mechanism that further boosts performance. Our study empirically indicates the potential to make MEO a golden standard efficient architecture within the NLP community. 

\section{Limitations}
\label{:sec:Limitations}
Despite the progress we have made, there are still limitations in our work. While our architecture for the mixture of experts demonstrates improved efficiency, there is a need for further exploration in terms of its deployment. Specifically, determining the optimal number of experts in specific layers and selecting different levels of MoEs require additional investigation. We believe that with the implementation of efficient deployment strategies, our method has the potential to become even more competitive.

\section*{Acknowledgements}
We are grateful to the anonymous EMNLP reviewers and the area chair for their insightful comments and suggestions.

\section*{Ethics Statement}
We take ethical considerations seriously and strictly adhere to the EMNLP Ethics Policy. This paper focuses on the higher efficiency of dynamic networks, e.g., the mixture of experts. Both the datasets and models used in this paper are publicly available and have been widely adopted by researchers. We ensure that the findings and conclusions of this paper are reported accurately and objectively.

\bibliography{anthology,custom}

\begin{thebibliography}{37}
\expandafter\ifx\csname natexlab\endcsname\relax\def\natexlab#1{#1}\fi

\bibitem[{Artetxe et~al.(2022)Artetxe, Bhosale, Goyal, Mihaylov, Ott, Shleifer, Lin, Du, Iyer, Pasunuru, Anantharaman, Li, Chen, Akin, Baines, Martin, Zhou, Koura, O'Horo, Wang, Zettlemoyer, Diab, Kozareva, and Stoyanov}]{DBLP:conf/emnlp/ArtetxeBGMOSLDI22}
Mikel Artetxe, Shruti Bhosale, Naman Goyal, Todor Mihaylov, Myle Ott, Sam Shleifer, Xi~Victoria Lin, Jingfei Du, Srinivasan Iyer, Ramakanth Pasunuru, Giridharan Anantharaman, Xian Li, Shuohui Chen, Halil Akin, Mandeep Baines, Louis Martin, Xing Zhou, Punit~Singh Koura, Brian O'Horo, Jeffrey Wang, Luke Zettlemoyer, Mona~T. Diab, Zornitsa Kozareva, and Veselin Stoyanov. 2022.
\newblock \href {https://aclanthology.org/2022.emnlp-main.804} {Efficient large scale language modeling with mixtures of experts}.
\newblock In \emph{EMNLP}.

\bibitem[{Bhargava et~al.(2021)Bhargava, Drozd, and Rogers}]{bhargava-2021-generalization}
Prajjwal Bhargava, Aleksandr Drozd, and Anna Rogers. 2021.
\newblock \href {https://doi.org/10.18653/v1/2021.insights-1.18} {Generalization in {NLI}: Ways (not) to go beyond simple heuristics}.
\newblock In \emph{the Second Workshop on Insights from Negative Results in NLP}.

\bibitem[{Brown et~al.(2020)Brown, Mann, Ryder, Subbiah, Kaplan, Dhariwal, Neelakantan, Shyam, Sastry, Askell, Agarwal, Herbert{-}Voss, Krueger, Henighan, Child, Ramesh, Ziegler, Wu, Winter, Hesse, Chen, Sigler, Litwin, Gray, Chess, Clark, Berner, McCandlish, Radford, Sutskever, and Amodei}]{DBLP:conf/nips/BrownMRSKDNSSAA20}
Tom~B. Brown, Benjamin Mann, Nick Ryder, Melanie Subbiah, Jared Kaplan, Prafulla Dhariwal, Arvind Neelakantan, Pranav Shyam, Girish Sastry, Amanda Askell, Sandhini Agarwal, Ariel Herbert{-}Voss, Gretchen Krueger, Tom Henighan, Rewon Child, Aditya Ramesh, Daniel~M. Ziegler, Jeffrey Wu, Clemens Winter, Christopher Hesse, Mark Chen, Eric Sigler, Mateusz Litwin, Scott Gray, Benjamin Chess, Jack Clark, Christopher Berner, Sam McCandlish, Alec Radford, Ilya Sutskever, and Dario Amodei. 2020.
\newblock \href {https://proceedings.neurips.cc/paper/2020/hash/1457c0d6bfcb4967418bfb8ac142f64a-Abstract.html} {Language models are few-shot learners}.
\newblock In \emph{NeurIPS}.

\bibitem[{Chen et~al.(2020)Chen, Dai, Liu, Chen, Yuan, and Liu}]{DBLP:conf/cvpr/ChenDLCYL20}
Yinpeng Chen, Xiyang Dai, Mengchen Liu, Dongdong Chen, Lu~Yuan, and Zicheng Liu. 2020.
\newblock \href {https://doi.org/10.1109/CVPR42600.2020.01104} {Dynamic convolution: Attention over convolution kernels}.
\newblock In \emph{CVPR}.

\bibitem[{Devlin et~al.(2019)Devlin, Chang, Lee, and Toutanova}]{devlin2018bert}
Jacob Devlin, Ming-Wei Chang, Kenton Lee, and Kristina Toutanova. 2019.
\newblock \href {https://doi.org/10.18653/v1/n19-1423} {Bert: Pre-training of deep bidirectional transformers for language understanding}.
\newblock In \emph{NAACL}.

\bibitem[{Dodge et~al.(2020)Dodge, Ilharco, Schwartz, Farhadi, Hajishirzi, and Smith}]{dodge2020fine}
Jesse Dodge, Gabriel Ilharco, Roy Schwartz, Ali Farhadi, Hannaneh Hajishirzi, and Noah Smith. 2020.
\newblock \href {https://arxiv.org/abs/2002.06305} {Fine-tuning pretrained language models: Weight initializations, data orders, and early stopping}.
\newblock \emph{arXiv preprint}.

\bibitem[{Du et~al.(2022)Du, Huang, Dai, Tong, Lepikhin, Xu, Krikun, Zhou, Yu, Firat, Zoph, Fedus, Bosma, Zhou, Wang, Wang, Webster, Pellat, Robinson, Meier{-}Hellstern, Duke, Dixon, Zhang, Le, Wu, Chen, and Cui}]{DBLP:conf/icml/DuHDTLXKZYFZFBZ22}
Nan Du, Yanping Huang, Andrew~M. Dai, Simon Tong, Dmitry Lepikhin, Yuanzhong Xu, Maxim Krikun, Yanqi Zhou, Adams~Wei Yu, Orhan Firat, Barret Zoph, Liam Fedus, Maarten~P. Bosma, Zongwei Zhou, Tao Wang, Yu~Emma Wang, Kellie Webster, Marie Pellat, Kevin Robinson, Kathleen~S. Meier{-}Hellstern, Toju Duke, Lucas Dixon, Kun Zhang, Quoc~V. Le, Yonghui Wu, Zhifeng Chen, and Claire Cui. 2022.
\newblock \href {https://proceedings.mlr.press/v162/du22c.html} {Glam: Efficient scaling of language models with mixture-of-experts}.
\newblock In \emph{ICML}.

\bibitem[{Fedus et~al.(2021)Fedus, Zoph, and Shazeer}]{fedus2021switch}
William Fedus, Barret Zoph, and Noam Shazeer. 2021.
\newblock \href {https://www.jmlr.org/papers/volume23/21-0998/21-0998.pdf} {Switch transformers: Scaling to trillion parameter models with simple and efficient sparsity}.
\newblock \emph{J. Mach. Learn. Res}.

\bibitem[{Gao et~al.(2022)Gao, Liu, Zhao, Lu, and Wen}]{gao-2022-parameter}
Ze-Feng Gao, Peiyu Liu, Wayne~Xin Zhao, Zhong-Yi Lu, and Ji-Rong Wen. 2022.
\newblock \href {https://aclanthology.org/2022.coling-1.288} {Parameter-efficient mixture-of-experts architecture for pre-trained language models}.
\newblock In \emph{COLING}.

\bibitem[{He et~al.(2023{\natexlab{a}})He, Ding, Dong, Liu, Yu, and Tao}]{he-etal-2023-pad}
Shwai He, Liang Ding, Daize Dong, Boan Liu, Fuqiang Yu, and Dacheng Tao. 2023{\natexlab{a}}.
\newblock \href {https://aclanthology.org/2023.acl-long.803} {{PAD}-net: An efficient framework for dynamic networks}.
\newblock In \emph{ACL}.

\bibitem[{He et~al.(2023{\natexlab{b}})He, Fan, Ding, Shen, Zhou, and Tao}]{he2023mera}
Shwai He, Run-Ze Fan, Liang Ding, Li~Shen, Tianyi Zhou, and Dacheng Tao. 2023{\natexlab{b}}.
\newblock \href {https://arxiv.org/abs/2308.15982} {Mera: Merging pretrained adapters for few-shot learning}.
\newblock \emph{arXiv preprint}.

\bibitem[{Houlsby et~al.(2019)Houlsby, Giurgiu, Jastrzebski, Morrone, De~Laroussilhe, Gesmundo, Attariyan, and Gelly}]{houlsby2019parameter}
Neil Houlsby, Andrei Giurgiu, Stanislaw Jastrzebski, Bruna Morrone, Quentin De~Laroussilhe, Andrea Gesmundo, Mona Attariyan, and Sylvain Gelly. 2019.
\newblock \href {http://proceedings.mlr.press/v97/houlsby19a.html} {Parameter-efficient transfer learning for nlp}.
\newblock In \emph{ICML}.

\bibitem[{Jacobs et~al.(1991)Jacobs, Jordan, Nowlan, and Hinton}]{DBLP:journals/neco/JacobsJNH91}
Robert~A. Jacobs, Michael~I. Jordan, Steven~J. Nowlan, and Geoffrey~E. Hinton. 1991.
\newblock \href {https://doi.org/10.1162/neco.1991.3.1.79} {Adaptive mixtures of local experts}.
\newblock \emph{Neural Comput.}

\bibitem[{Jordan and Jacobs(1994)}]{DBLP:journals/neco/JordanJ94}
Michael~I. Jordan and Robert~A. Jacobs. 1994.
\newblock \href {https://doi.org/10.1162/neco.1994.6.2.181} {Hierarchical mixtures of experts and the {EM} algorithm}.
\newblock \emph{Neural Comput.}

\bibitem[{Kingma and Ba(2015)}]{kingma2014adam}
Diederik~P Kingma and Jimmy Ba. 2015.
\newblock \href {http://arxiv.org/abs/1412.6980} {Adam: A method for stochastic optimization}.
\newblock In \emph{ICLR}.

\bibitem[{Kudugunta et~al.(2021)Kudugunta, Huang, Bapna, Krikun, Lepikhin, Luong, and Firat}]{DBLP:conf/emnlp/KuduguntaHBKLLF21}
Sneha Kudugunta, Yanping Huang, Ankur Bapna, Maxim Krikun, Dmitry Lepikhin, Minh{-}Thang Luong, and Orhan Firat. 2021.
\newblock \href {https://doi.org/10.18653/v1/2021.findings-emnlp.304} {Beyond distillation: Task-level mixture-of-experts for efficient inference}.
\newblock In \emph{EMNLP}.

\bibitem[{Lee et~al.(2020)Lee, Cho, and Kang}]{lee2019mixout}
Cheolhyoung Lee, Kyunghyun Cho, and Wanmo Kang. 2020.
\newblock \href {https://openreview.net/forum?id=HkgaETNtDB} {Mixout: Effective regularization to finetune large-scale pretrained language models}.
\newblock In \emph{ICLR}.

\bibitem[{Lewis et~al.(2020)Lewis, Liu, Goyal, Ghazvininejad, Mohamed, Levy, Stoyanov, and Zettlemoyer}]{lewisetal2020-bart}
Mike Lewis, Yinhan Liu, Naman Goyal, Marjan Ghazvininejad, Abdelrahman Mohamed, Omer Levy, Veselin Stoyanov, and Luke Zettlemoyer. 2020.
\newblock \href {https://doi.org/10.18653/v1/2020.acl-main.703} {{BART}: Denoising sequence-to-sequence pre-training for natural language generation, translation, and comprehension}.
\newblock In \emph{ACL}.

\bibitem[{Li et~al.(2021)Li, Zhou, and Yao}]{li2021omni}
Chao Li, Aojun Zhou, and Anbang Yao. 2021.
\newblock \href {https://openreview.net/forum?id=DmpCfq6Mg39} {Omni-dimensional dynamic convolution}.
\newblock In \emph{ICLR}.

\bibitem[{Li et~al.(2023)Li, Peng, Zhang, Ding, Hu, and Shen}]{li2023deep}
Weishi Li, Yong Peng, Miao Zhang, Liang Ding, Han Hu, and Li~Shen. 2023.
\newblock \href {https://arxiv.org/abs/2309.15698} {Deep model fusion: A survey}.
\newblock \emph{arXiv preprint}.

\bibitem[{Merity et~al.(2016)Merity, Xiong, Bradbury, and Socher}]{merity2016pointer}
Stephen Merity, Caiming Xiong, James Bradbury, and Richard Socher. 2016.
\newblock \href {http://arxiv.org/abs/1609.07843} {Pointer sentinel mixture models}.

\bibitem[{Mustafa et~al.(2022)Mustafa, Riquelme, Puigcerver, Jenatton, and Houlsby}]{DBLP:conf/nips/MustafaRPJH22}
Basil Mustafa, Carlos Riquelme, Joan Puigcerver, Rodolphe Jenatton, and Neil Houlsby. 2022.
\newblock \href {http://papers.nips.cc/paper\_files/paper/2022/hash/3e67e84abf900bb2c7cbd5759bfce62d-Abstract-Conference.html} {Multimodal contrastive learning with limoe: the language-image mixture of experts}.
\newblock In \emph{NeurIPS}.

\bibitem[{Narayan et~al.(2018)Narayan, Cohen, and Lapata}]{narayanetal-2018-dont}
Shashi Narayan, Shay~B. Cohen, and Mirella Lapata. 2018.
\newblock \href {https://doi.org/10.18653/v1/D18-1206} {Don{'}t give me the details, just the summary! topic-aware convolutional neural networks for extreme summarization}.
\newblock In \emph{EMNLP}.

\bibitem[{OpenAI(2023)}]{DBLP:journals/corr/abs-2303-08774}
OpenAI. 2023.
\newblock \href {https://doi.org/10.48550/arXiv.2303.08774} {{GPT-4} technical report}.
\newblock \emph{arXiv preprint}.

\bibitem[{Pfeiffer et~al.(2021)Pfeiffer, Kamath, R{\"u}ckl{\'e}, Cho, and Gurevych}]{pfeiffer2020adapterfusion}
Jonas Pfeiffer, Aishwarya Kamath, Andreas R{\"u}ckl{\'e}, Kyunghyun Cho, and Iryna Gurevych. 2021.
\newblock \href {https://doi.org/10.18653/v1/2021.eacl-main.39} {Adapterfusion: Non-destructive task composition for transfer learning}.
\newblock In \emph{EACL}.

\bibitem[{Phang et~al.(2018)Phang, F{\'e}vry, and Bowman}]{phang2018sentence}
Jason Phang, Thibault F{\'e}vry, and Samuel~R Bowman. 2018.
\newblock \href {http://arxiv.org/abs/1811.01088} {Sentence encoders on stilts: Supplementary training on intermediate labeled-data tasks}.
\newblock \emph{arXiv preprint}.

\bibitem[{Radford et~al.(2019)Radford, Wu, Child, Luan, Amodei, and Sutskever}]{radford2019language}
Alec Radford, Jeff Wu, Rewon Child, David Luan, Dario Amodei, and Ilya Sutskever. 2019.
\newblock \href {https://insightcivic.s3.us-east-1.amazonaws.com/language-models.pdf} {Language models are unsupervised multitask learners}.

\bibitem[{Raffel et~al.(2020)Raffel, Shazeer, Roberts, Lee, Narang, Matena, Zhou, Li, and Liu}]{2020t5}
Colin Raffel, Noam Shazeer, Adam Roberts, Katherine Lee, Sharan Narang, Michael Matena, Yanqi Zhou, Wei Li, and Peter~J. Liu. 2020.
\newblock \href {http://jmlr.org/papers/v21/20-074.html} {Exploring the limits of transfer learning with a unified text-to-text transformer}.
\newblock \emph{J. Mach. Learn. Res}.

\bibitem[{Rajpurkar et~al.(2016)Rajpurkar, Zhang, Lopyrev, and Liang}]{DBLP:conf/emnlp/RajpurkarZLL16}
Pranav Rajpurkar, Jian Zhang, Konstantin Lopyrev, and Percy Liang. 2016.
\newblock \href {https://doi.org/10.18653/v1/d16-1264} {{SQuAD}: 100, 000+ questions for machine comprehension of text}.
\newblock In \emph{EMNLP}.

\bibitem[{Shazeer et~al.(2017)Shazeer, Mirhoseini, Maziarz, Davis, Le, Hinton, and Dean}]{DBLP:conf/iclr/ShazeerMMDLHD17}
Noam Shazeer, Azalia Mirhoseini, Krzysztof Maziarz, Andy Davis, Quoc~V. Le, Geoffrey~E. Hinton, and Jeff Dean. 2017.
\newblock \href {https://openreview.net/forum?id=B1ckMDqlg} {Outrageously large neural networks: The sparsely-gated mixture-of-experts layer}.
\newblock In \emph{ICLR}.

\bibitem[{Wang et~al.(2019)Wang, Singh, Michael, Hill, Levy, and Bowman}]{wang2018glue}
Alex Wang, Amanpreet Singh, Julian Michael, Felix Hill, Omer Levy, and Samuel~R Bowman. 2019.
\newblock \href {https://openreview.net/forum?id=rJ4km2R5t7} {{GLUE:} {A} multi-task benchmark and analysis platform for natural language understanding}.
\newblock In \emph{ICLR}.

\bibitem[{Wang et~al.(2022)Wang, Zhang, Sun, and Meng}]{DBLP:conf/emnlp/wangcort22}
Yequan Wang, Hengran Zhang, Aixin Sun, and Xuying Meng. 2022.
\newblock \href {https://doi.org/doi = "10.18653/v1/2022.findings-emnlp.524"} {Cort: A new baseline for comparative opinion classification by dual prompts}.
\newblock In \emph{EMNLP}.

\bibitem[{Yang et~al.(2019)Yang, Bender, Le, and Ngiam}]{DBLP:conf/nips/YangBLN19}
Brandon Yang, Gabriel Bender, Quoc~V. Le, and Jiquan Ngiam. 2019.
\newblock \href {https://proceedings.neurips.cc/paper/2019/hash/f2201f5191c4e92cc5af043eebfd0946-Abstract.html} {Condconv: Conditionally parameterized convolutions for efficient inference}.
\newblock In \emph{NeurIPS}.

\bibitem[{Ye et~al.(2022)Ye, Zha, and Ren}]{ye-2022-eliciting}
Qinyuan Ye, Juan Zha, and Xiang Ren. 2022.
\newblock \href {https://aclanthology.org/2022.findings-emnlp.189} {Eliciting and understanding cross-task skills with task-level mixture-of-experts}.
\newblock In \emph{EMNLP}.

\bibitem[{Zhong et~al.(2022{\natexlab{a}})Zhong, Ding, Liu, Du, and Tao}]{zhong2022e2s2}
Qihuang Zhong, Liang Ding, Juhua Liu, Bo~Du, and Dacheng Tao. 2022{\natexlab{a}}.
\newblock \href {https://arxiv.org/abs/2205.14912} {E2s2: Encoding-enhanced sequence-to-sequence pretraining for language understanding and generation}.
\newblock \emph{arXiv preprint}.

\bibitem[{Zhong et~al.(2022{\natexlab{b}})Zhong, Ding, Shen, Mi, Liu, Du, and Tao}]{zhong2022improving}
Qihuang Zhong, Liang Ding, Li~Shen, Peng Mi, Juhua Liu, Bo~Du, and Dacheng Tao. 2022{\natexlab{b}}.
\newblock \href {https://aclanthology.org/2022.findings-emnlp.300/} {Improving sharpness-aware minimization with fisher mask for better generalization on language models}.
\newblock In \emph{EMNLP}.

\bibitem[{Zhu et~al.(2022)Zhu, Zhu, Wang, Wang, Li, Wang, and Dai}]{DBLP:conf/nips/ZhuZWWLWD22}
Jinguo Zhu, Xizhou Zhu, Wenhai Wang, Xiaohua Wang, Hongsheng Li, Xiaogang Wang, and Jifeng Dai. 2022.
\newblock \href {http://papers.nips.cc/paper\_files/paper/2022/hash/11fc8c98b46d4cbdfe8157267228f7d7-Abstract-Conference.html} {Uni-perceiver-moe: Learning sparse generalist models with conditional moes}.
\newblock In \emph{NeurIPS}.

\end{thebibliography}
\bibliographystyle{acl_natbib}

\end{document}